\documentclass[11pt]{article}

\usepackage[a4paper,margin=1in]{geometry}
\usepackage{graphicx}
\usepackage{amsmath,amssymb}
\usepackage{booktabs}
\usepackage{multirow}
\usepackage{array}
\usepackage{float}
\usepackage{placeins}

\usepackage{xcolor}
\usepackage{caption}
\usepackage{subcaption}
\usepackage{microtype}
\usepackage[numbers,sort&compress]{natbib}
\usepackage{hyperref}
\hypersetup{
    colorlinks=true,
    linkcolor=blue,
    citecolor=blue,
    urlcolor=blue
}

\title{A Task-Driven Evaluation of UAV Detection and Tracking under Synthetic Fog}

\author{
Amir Pouladi$^{1,2,*}$,
Vesal Ahsani$^{1}$,
Haijun Li$^{1}$,
Homayoun Najjaran$^{2}$,
and Afzal Suleman$^{1}$\\[0.5em]
\small
\begin{tabular}{c}
$^{1}$Center for Aerospace Research (CfAR), University of Victoria,\\
Victoria, BC, Canada\\
$^{2}$Advanced Control and Intelligent Systems (ACIS) Lab, University of Victoria,\\
Victoria, BC, Canada\\[0.3em]
$^{*}$Corresponding author: \texttt{apouladi@uvic.ca}
\end{tabular}
}

\date{}

\begin{document}

\maketitle

\begin{abstract}
Fog severely degrades the visibility of small unmanned aerial vehicles (UAVs) in sky-dominant, long-range imagery, reducing the reliability of downstream detection and tracking. This paper presents a task-driven evaluation framework that links depth-aware synthetic fog generation, image restoration, object detection, and tracking within a unified pipeline. Given the practical difficulty of collecting and annotating foggy UAV scenes, synthetic fog is generated from real clear-weather outdoor images containing UAV targets using monocular depth estimation and the atmospheric scattering model. Representative restoration methods from classical, convolutional neural network (CNN)-based, and transformer-based families are first compared, after which the selected restoration model is integrated into the downstream perception pipeline. Detection is evaluated under both clean-only and fog-inclusive training regimes using multiple detector variants, while tracking-by-detection is assessed on clean, foggy, and restored video sequences. Beyond image-level restoration metrics, the study evaluates how fog and restoration affect detection robustness and tracking performance. The results show that fog substantially degrades both detection and tracking, primarily through increased missed detections. Fog-inclusive training provides the most consistent improvement in robustness, whereas test-time restoration is most beneficial when the detector has been trained only on clean imagery. These findings show that restoration quality does not necessarily translate into proportional gains in downstream perception and therefore should be evaluated jointly with detection and tracking performance.
\end{abstract}


\section{Introduction}

Reliable detection and tracking of small unmanned aerial vehicles (UAVs) in open airspace is a core capability for airborne situational awareness, and sense-and-avoid perception. Under fog, image formation is dominated by depth-dependent attenuation and airlight, which jointly reduce target--background contrast and suppress high-frequency details that are critical for recognizing small, distant objects. Classical atmospheric visibility models describe this degradation through an exponential transmission term and additive path radiance, linking perceived contrast loss to scene depth and an extinction coefficient \cite{mccartney1976optics}. In practice, these effects can cause severe performance drops for detectors and trackers trained primarily on clear imagery.

Single-image dehazing has been extensively studied as a low-level restoration problem, with substantial progress from hand-crafted priors to learning-based methods \cite{song2023vision}. Large-scale benchmarks such as RESIDE have also emphasized that improvements in image-level criteria do not necessarily imply improvements for downstream vision tasks \cite{li2018benchmarking}. This gap is especially important for airborne UAV perception in sky-dominant, long-range imagery, where targets often occupy only a few pixels, and local edge fidelity and background smoothness can strongly affect detector confidence and the temporal stability required for tracking-by-detection.

At the same time, the UAV detection and tracking community has produced challenging benchmarks that highlight the difficulty of small-target perception, including Anti-UAV~\cite{jiang2021anti} and DUT Anti-UAV~\cite{zhao2022vision}, which provide annotated sequences for detection and tracking evaluation. Recent multi-modal datasets such as MMAUD~\cite{yuan2024mmaud} further underscore the operational relevance of adverse conditions for anti-UAV sensing and algorithm development. However, a systematic, task-level study that connects depth-aware fog synthesis, image dehazing, and downstream UAV detection and tracking within a controlled protocol remains limited, particularly for sky-dominant scenes where fog effects are strongly depth-driven.

This paper addresses this gap by presenting a task-driven, end-to-end evaluation pipeline that couples synthetic fog generation, image restoration, object detection, and object tracking. The goal is not to propose a new restoration model, detector, or tracker, but to provide a rigorous experimental protocol to answer two practical research questions:

\begin{itemize}
    \item When fog is present at inference, to what extent can dehazing recover UAV detection and tracking performance relative to foggy input, and how does this depend on fog density?
    \item How does fog-inclusive detector training compare with test-time restoration, and under what conditions does training with synthetic fog reduce reliance on restoration for robust detection and tracking?
\end{itemize}

\subsection{Contributions}
The main contributions of this work are summarized as follows:

\begin{itemize}
    
    \item A task-driven evaluation framework is presented for UAV detection and tracking under adverse visibility, with emphasis on downstream perception performance rather than image-level restoration quality alone.
    
    \item A depth-aware synthetic fog generation pipeline is developed for real sky-dominant UAV imagery and used to construct controlled fog-degraded benchmark variants of existing UAV datasets across multiple fog severities. The generated foggy datasets are publicly available.

    \item A unified experimental study is conducted across image restoration, fog-inclusive detector training, and tracking-by-detection to analyze how weather degradation and restoration affect detection robustness and temporal tracking performance.
\end{itemize}

\section{Related Work}
\label{sec:related_work}
The related work most relevant to this study falls into three areas, namely synthetic fog generation, image dehazing, and detection and tracking under fog. 

\subsection{Synthetic Fog Generation for Adverse-Weather Perception}

Synthetic fog generation has become an important strategy for studying adverse-weather perception because paired clear/foggy data with reliable annotations are difficult to acquire in real outdoor environments. Most existing methods are ultimately rooted in physics-based atmospheric image formation, where scene radiance is attenuated along the line of sight and mixed with airlight. Foundational work by Narasimhan and Nayar~\cite{narasimhan2002vision} established the imaging principles of atmospheric scattering and provided the physical basis that later computer-vision methods used for fog and haze synthesis as well as restoration.

In autonomous driving, the dominant line of work has relied on image-level fog synthesis guided by scene depth. An important contribution is Foggy Cityscapes by Sakaridis \emph{et al.}~\cite{sakaridis2018semantic}, who introduced a pipeline for adding synthetic fog to real clear-weather urban scenes using incomplete depth information, thereby enabling supervised training for foggy-scene understanding. This direction was extended by curriculum-based adaptation strategies that combined synthetic fog with unlabeled real foggy data, improving robustness under denser and more realistic fog conditions~\cite{sakaridis2018model}. More recent driving-oriented studies have refined this paradigm by estimating depth directly from monocular RGB images and then applying an atmospheric scattering model. For example, Nie \emph{et al.}~\cite{nie2022foggy} synthesized a foggy lane-detection dataset from monocular images using monocular depth prediction and atmospheric scattering, while Tang \emph{et al.}~\cite{tang2024foggy} further incorporated self-supervised monocular depth estimation, scale recovery, transmittance-map generation, sky-region handling, and atmospheric-light estimation to obtain more realistic fog simulation for traffic-image augmentation. Beyond classical image-space atmospheric scattering, newer work has moved toward more comprehensive physical simulation. SynFog~\cite{xie2024synfog}, for instance, models the full foggy imaging chain in an end-to-end physically based manner and reports improved transfer to real-world foggy perception compared with simpler atmospheric-scattering or rendering-engine baselines. In parallel, fog synthesis has also been extended beyond RGB imagery. Hahner \emph{et al.}~\cite{hahner2021fog} proposed a physically valid fog simulation method for real LiDAR point clouds and showed that synthetic fog can improve 3D object detection robustness on real foggy driving data.

Compared with autonomous driving, the UAV literature on synthetic fog generation is still more limited and can be broadly divided into airborne-target detection and drone-view air-to-ground object detection. In the airborne-target setting, Zheng et al.~\cite{zheng2025foggy} proposed Foggy Drone Teacher (FDT), a Mean Teacher-based domain-adaptive framework for UAV detection under foggy conditions, and introduced the Foggy Drone Dataset (FDD) generated using the atmospheric scattering model. While their work demonstrates the benefit of domain adaptation for foggy UAV detection, the dataset appears to use a single fog configuration, and the paper does not clearly describe the depth-estimation or normalization process used for fog synthesis, which may limit the controllability of fog severity and depth-dependent visibility evaluation. In drone-view air-to-ground detection, Xi \emph{et al.}~\cite{xi2022fifonet} generated a synthetic VisDrone\_Foggy dataset from VisDrone2019~\cite{wen2021detection} using different fog densities and used it to evaluate fine-grained detection in drone imagery. Fang \emph{et al.}~\cite{fang2023multi} addressed foggy UAV perception through a joint dehazing-and-detection framework, reflecting a common trend in the UAV literature in which synthetic fog is used primarily as a training or evaluation perturbation rather than as a rigorously validated environmental model. More recently, HazyDet~\cite{feng2024hazydet} introduced a dedicated drone-view benchmark containing both naturally hazy imagery and synthetically degraded scenes. However, recent reviews of UAV perception under adversity still indicate that fog-robust airborne perception remains underdeveloped relative to road-scene perception, especially for safety-critical onboard obstacle or aircraft detection~\cite{randieri2025aerial}.

\subsection{Image dehazing Methods}
Single-image dehazing aims to recover clean scene radiance from a haze or fog degraded observation. The literature can be organized into four broad groups, classical prior-based and optimization-based methods, CNN-based methods, transformer-based methods, and recent diffusion-based methods. Systematic evaluation of dehazing methods was accelerated by RESIDE dataset~\cite{li2018benchmarking}, which standardized paired synthetic evaluation as well as real-image testing for dehazing methods. Early dehazing methods relied on handcrafted priors and explicit assumptions on haze formation. Tan enhanced visibility by maximizing local contrast with smoothness constraints on airlight~\cite{tan2008visibility}. Fattal exploited statistical independence between shading and transmission for single-image dehazing~\cite{fattal2008single}. A major milestone is the dark channel prior (DCP) of He \emph{et al.}, which estimates transmission using patch-wise dark-channel statistics of haze-free images~\cite{he2010single}. Meng \emph{et al.} introduced boundary constraint and contextual regularization to improve transmission estimation and reduce edge artifacts~\cite{meng2013efficient}. Zhu \emph{et al.} proposed the color attenuation prior (CAP), which learns a linear relation between color attenuation and scene depth, providing a computationally efficient formulation~\cite{zhu2015fast}. Berman \emph{et al.} later developed non-local and haze-line formulations that exploit color-line structure in RGB space for deterministic dehazing without training~\cite{berman2016non,berman2018single}. These classical methods remain attractive because they are interpretable and can be computationally lightweight. However, they are often sensitive to their underlying assumptions, such as locally dark patches or homogeneous scattering. 

CNN-based dehazing first emerged through networks that estimate transmission maps and later evolved into end-to-end image restoration. DehazeNet is one of the earliest end-to-end CNN dehazing models, learning haze-relevant features to estimate transmission~\cite{cai2016dehazenet}. Ren \emph{et al.} introduced a multi-scale CNN to improve receptive-field coverage and transmission estimation stability~\cite{ren2016single}. AOD-Net reformulated the atmospheric scattering model so that restoration could be performed directly in an end-to-end network while remaining lightweight and suitable for integration into higher-level vision tasks~\cite{li2017aod}. Attention and multi-scale designs then became prominent, including GridDehazeNet~\cite{liu2019griddehazenet}, MSBDN~\cite{dong2020multi}, and FFA-Net~\cite{qin2020ffa}. More recent CNN approaches increasingly emphasize compactness, robustness to real-haze domain shift, and more effective priors. AECR-Net introduced contrastive regularization for compact dehazing~\cite{wu2021contrastive}. C2PNet proposed curricular contrastive regularization to better align physical constraints with learned features~\cite{zheng2023curricular}. RIDCP introduced high-quality codebook priors to revitalize real-image dehazing~\cite{wu2023ridcp}. DEA-Net combined detail-enhanced convolution and content-guided attention to improve restoration while maintaining a compact design~\cite{chen2024dea}.

Transformers were introduced into dehazing to better model long-range dependencies and global haze distribution. DeHamer used transmission-aware 3D position embeddings to inject physically meaningful cues into a transformer architecture~\cite{guo2022image}. DehazeFormer proposed a dehazing-specialized transformer architecture and showed that a compact transformer can outperform strong CNN baselines while remaining computationally efficient~\cite{song2023vision}. More recent hybrid architectures explicitly combine CNN locality with transformer global context, such as the interaction-guided two-branch design of Liu \emph{et al.}~\cite{liu2024interaction}.

Diffusion models have recently been explored for dehazing because of their strong generative priors, but their sampling complexity remains a major practical challenge. DehazeDDPM introduced a physics-aware diffusion formulation guided by the atmospheric scattering model~\cite{yu2023high}. DiffLI2D exploited the semantic latent space of pretrained diffusion models to build an efficient dehazing pipeline without full diffusion retraining~\cite{yang2024unleashing}. Recent work has also explored temporal-aware and latent-space diffusion variants to reduce inference cost while preserving restoration quality~\cite{liang2025efficient,liang2025latent}.

\subsection{Detection and Tracking under Fog}

Detection and tracking under fog have been studied mainly through three directions, namely synthetic adverse-weather data generation, adaptation of the perception model, and the use of temporal or multimodal cues. In autonomous-driving research, Sakaridis \emph{et al.} showed that synthetic fog can support fog-aware scene understanding and that image dehazing provides only limited gains for downstream object detection compared with learning on fog-aware data~\cite{sakaridis2018semantic,sakaridis2018model}. More recent work has reinforced this view. Gupta \emph{et al.} compared all-weather training, synthetic weather augmentation, and pre-detection image restoration for object detection in adverse weather, and found that training with adverse-weather data was generally more effective than applying restoration before detection~\cite{gupta2024robust}. In parallel, robustness-oriented detection methods have been proposed to reduce the performance gap between clear and degraded domains. Fu \emph{et al.} introduced auxiliary domain-guided adaptation for adverse-weather object detection~\cite{fu2024auxiliary}, while Ogino \emph{et al.} proposed ERUP-YOLO, which integrates differentiable image-adaptive preprocessing into the detector and reports improved robustness under fog and low-light conditions~\cite{ogino2025erup}.

In aerial perception, the literature remains more limited. Feng \emph{et al.} introduced HazyDet as a large-scale benchmark for drone-view object detection in hazy scenes and proposed a depth-conditioned detector to address the combined effects of haze and scale variation~\cite{feng2024hazydet}. For small-object detection in real adverse weather, van Lier \emph{et al.} evaluated a spatio-temporal detector on real-world adverse-weather data and showed that temporal modeling and haze-aware data design can improve robustness under degraded visibility~\cite{van2025evaluation}. These studies suggest that adverse-weather robustness is increasingly addressed through data design and temporal modeling rather than through restoration alone.

Tracking under fog is less explored than detection. Feng \emph{et al.} proposed a two-stage aerial tracking framework that combines image restoration with tracking and reported improved performance under foggy conditions~\cite{feng2024two}. In the ground-vehicle domain, Ogunrinde proposed an improved DeepSORT-based tracker for foggy weather, highlighting the impact of degraded visual features on association errors and track failures~\cite{ogunrinde2024improved}. Beyond restoration-based pipelines, multimodal perception has emerged as a robust alternative under degraded visibility. Bijelic \emph{et al.} showed that deep multimodal fusion can improve object detection in unseen adverse weather, including fog, by exploiting complementary sensing modalities~\cite{bijelic2020seeing}. Overall, the literature suggests that fog degrades downstream perception primarily by weakening object evidence and increasing missed detections, while the most effective remedies tend to come from robustness-oriented training, temporal modeling, or multimodal fusion rather than from image enhancement alone.

\section{Materials and Methods}

\subsection{Overall Pipeline}
\begin{figure}[!htbp]
\includegraphics[width=\textwidth,trim={0.7cm 0.9cm 0.7cm 0.8cm},clip]{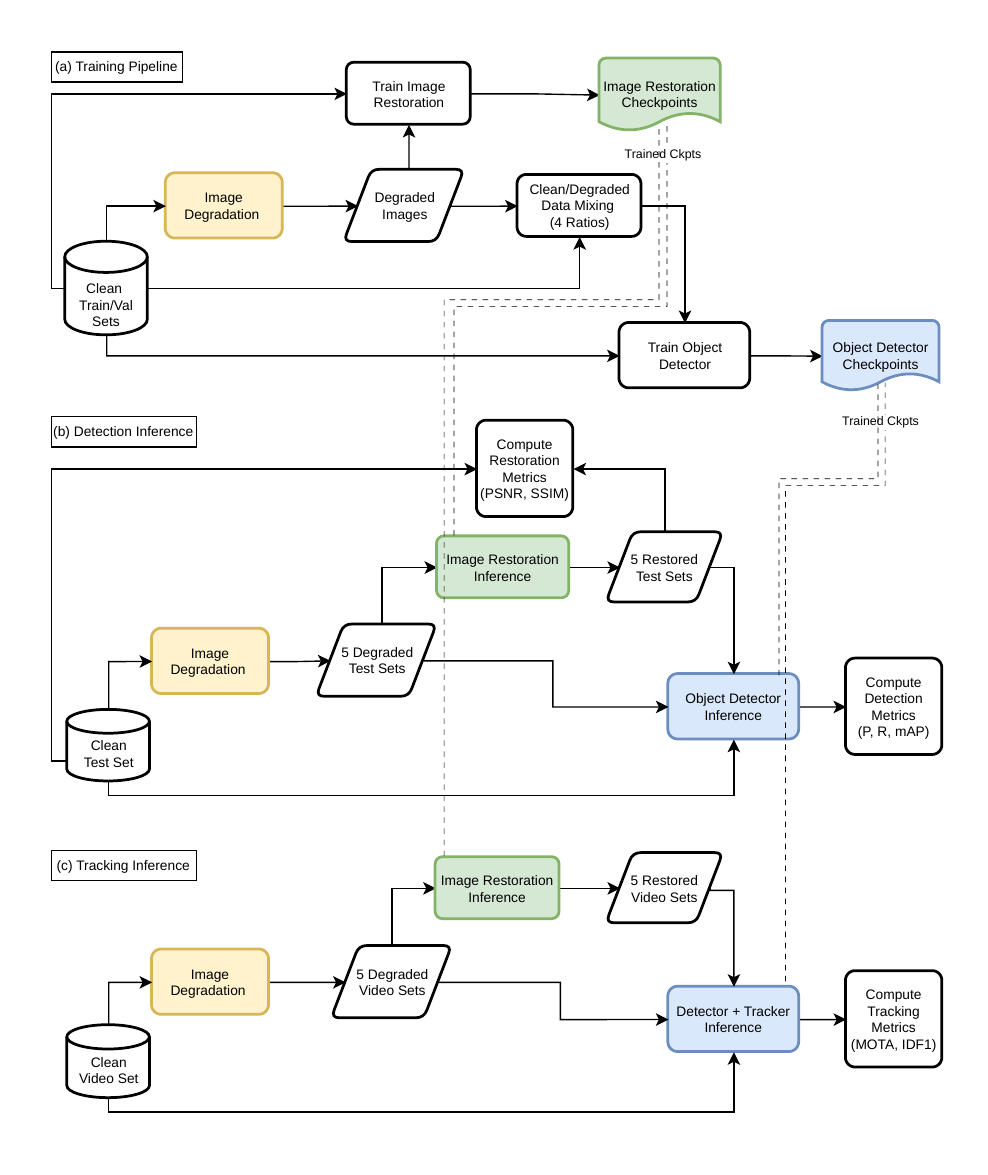}
\caption{End-to-end experimental pipeline used in this work.}
\label{fig:pipeline}
\end{figure}

As shown in Figure~\ref{fig:pipeline}, the proposed framework consists of three stages, namely training, detection inference, and tracking inference. In the training stage, clean training and validation images are first degraded using the proposed depth-aware fog synthesis pipeline to generate foggy counterparts. These degraded images are used in two ways. First, they are paired with the clean images to train the image restoration model. Second, they are mixed with the clean training data at four predefined ratios to train the object detectors under different fog-inclusive regimes. This stage produces both image restoration checkpoints and object detector checkpoints.

In the detection inference stage, the clean test set is degraded into five foggy test sets corresponding to the selected fog severities. The restoration model is then applied to these degraded test sets to generate five restored test sets. The trained detectors are evaluated on the clean, foggy, and restored test sets, and detection performance is quantified using precision, recall, and mAP. In parallel, restoration quality is evaluated using PSNR and SSIM.

In the tracking inference stage, the clean video set is similarly degraded to produce five foggy video sets, which are then restored to obtain five restored video sets. The selected detector and tracker are applied to the clean, foggy, and restored video sets in a tracking-by-detection setting. Tracking performance is then measured using MOTA and IDF1, allowing the effect of weather degradation and restoration to be examined at the temporal level.

\subsection{Synthetic Fog Generation}

Synthetic fog is generated from clean images using a depth-aware atmospheric scattering model (ASM). Monocular depth is first estimated using MiDaS~\cite{ranftl2021vision}, which provides a dense relative depth representation from a single RGB input. The foggy image is then synthesized according to
\begin{equation}
I(\mathbf{x}) = J(\mathbf{x})\,t(\mathbf{x}) + A\left(1-t(\mathbf{x})\right),
\label{eq:asm}
\end{equation}
where $J(\mathbf{x})$ denotes the clean scene radiance, $I(\mathbf{x})$ is the observed foggy image, $A$ is the global atmospheric light, and $t(\mathbf{x})$ is the transmission map defined as
\begin{equation}
t(\mathbf{x}) = \exp\left(-\beta\,d(\mathbf{x})\right),
\label{eq:transmission}
\end{equation}
with $d(\mathbf{x})$ representing depth and $\beta$ controlling the fog severity.

\begin{figure}[!htbp]
\includegraphics[width=\textwidth,trim={0.7cm 0.7cm 0.7cm 0.7cm},clip]{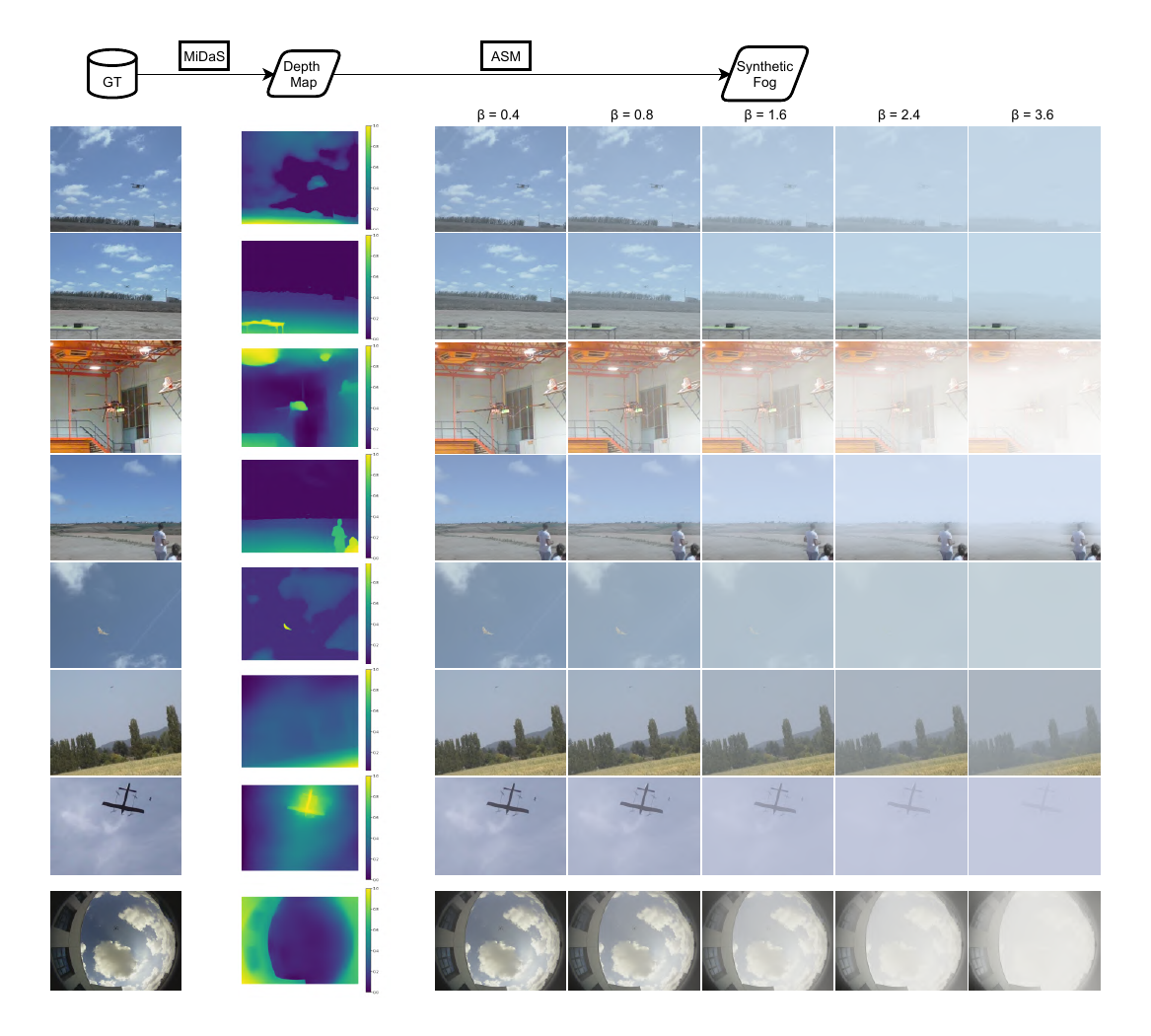}
%Left, Bottom, Right, Top
\caption{Depth-aware synthetic fog generation used in this study. Clean images are processed by MiDaS to obtain depth maps, which are then used in the atmospheric scattering model (ASM) to synthesize fog with increasing density levels $\beta$.}
\label{fig:degradation}
\end{figure}

Figure~\ref{fig:degradation} illustrates the overall pipeline. For each image, the raw MiDaS depth $d_{\mathrm{raw}}(\mathbf{x})$ is normalized using fixed split-level percentile bounds,
\[
d_{\mathrm{norm}}(\mathbf{x}) = \operatorname{clip}\left(\frac{d_{\mathrm{raw}}(\mathbf{x}) - p_{\mathrm{low}}}{p_{\mathrm{high}} - p_{\mathrm{low}}}, 0, 1\right),
\]
where $p_{\mathrm{low}}$ and $p_{\mathrm{high}}$ denote the lower and upper percentiles of the raw MiDaS depth distribution computed for each dataset split. This percentile-based normalization ensures consistent scaling across images while accounting for the scale ambiguity of monocular depth estimation. The normalized depth is then inverted so that larger values correspond to farther scene points,
\[
d(\mathbf{x}) = 1 - d_{\mathrm{norm}}(\mathbf{x}).
\]
The transmission map is computed via Eq.~(\ref{eq:transmission}) and further refined using guided filtering to suppress artifacts and better align depth discontinuities with image structures. Finally, the foggy image is obtained using Eq.~(\ref{eq:asm}), where $A$ is estimated from the brightest pixels of the input image.

The fog density parameter $\beta$ was varied over five levels, $\{0.4, 0.8, 1.6, 2.4, 3.6\}$, to generate a controlled progression from light to severe degradation. This follows established synthetic fog generation practices, where $\beta$ is used as the primary attenuation parameter and evaluated across multiple levels~\cite{sakaridis2018model,sakaridis2018semantic,nie2022foggy}. It is important to note that the numerical range of $\beta$ is not directly comparable across studies, as it depends on the underlying depth representation and scene scaling. For example, Foggy Cityscapes employs metric stereo depth and uses $\beta \in \{0, 0.005, 0.01, 0.02\}$~\cite{sakaridis2018model}, whereas Nie \emph{et al.} adopt monocular depth estimation and use $\beta \in \{2, 3, 4\}$~\cite{nie2022foggy}. These differences arise from variations in depth scaling and image formation pipelines rather than strictly different physical fog densities.

In this work, MiDaS provides relative (scale-ambiguous) depth estimates. Therefore, $\beta$ is interpreted as a dimensionless severity-control parameter governing the strength of depth-dependent attenuation. The selected values span a broad range of perceptually plausible fog severities for open-sky imagery while avoiding negligible degradation or near-complete visibility loss. Due to the exponential dependence of transmission on depth, distant regions are attenuated more strongly than near-field structures, yielding depth-varying fog that preserves scene geometry while progressively reducing contrast.

\subsection{Image Restoration Methods}

\begin{figure}[H]
\includegraphics[width=\textwidth,trim={0.8cm 0.55cm 0.7cm 0.6cm},clip]{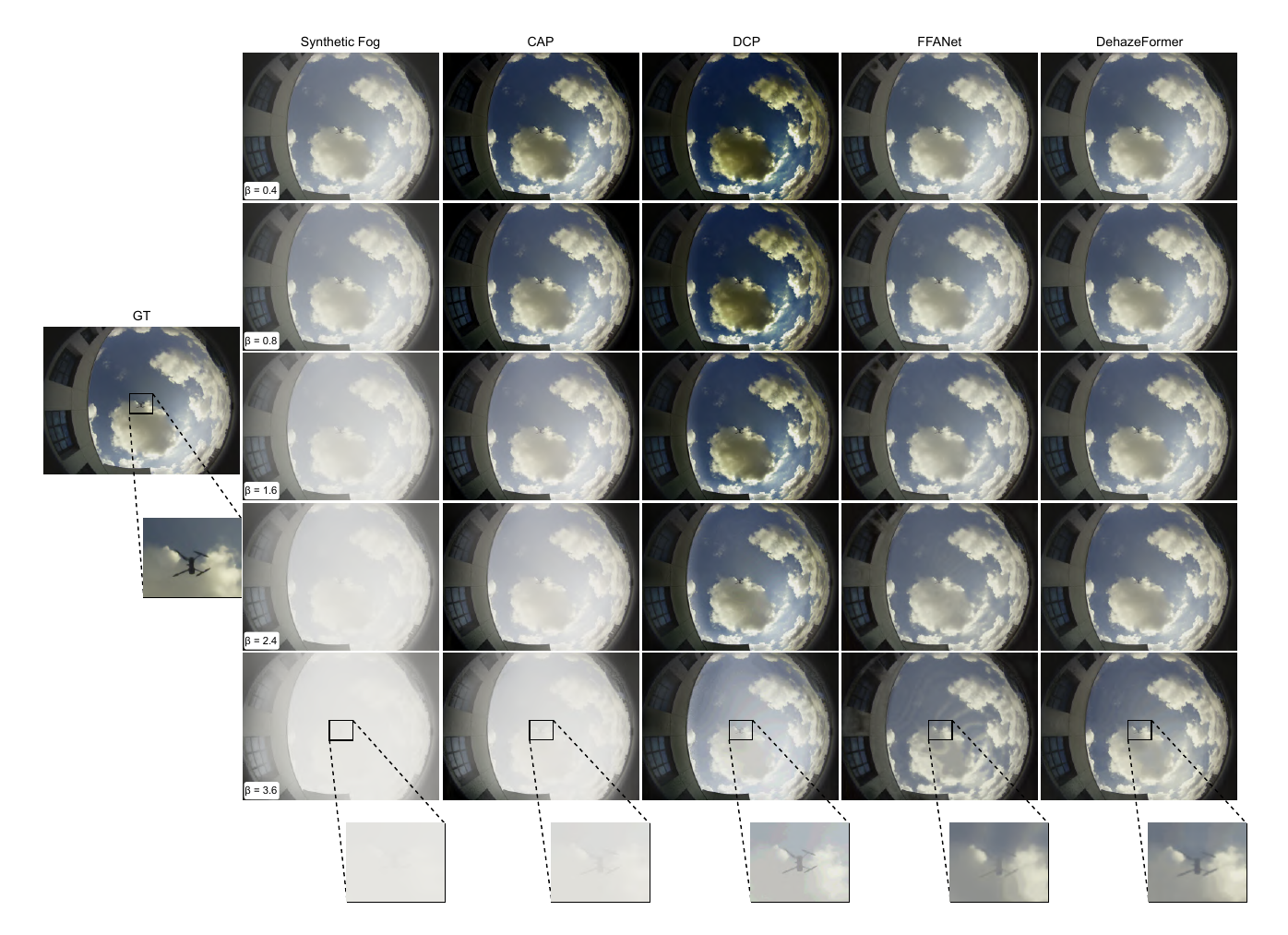}
\caption{Qualitative comparison of restoration methods across fog severities over MMAUD~\cite{yuan2024mmaud}. The columns show fog-degraded inputs and restored outputs from CAP, DCP, FFA-Net, and DehazeFormer for increasing fog densities. The zoomed region emphasizes restoration behavior around the UAV target, which is critical for downstream detection.}
\label{fig:restoration}
\end{figure}

The experiments consider representative image restoration methods from three practically relevant families: classical prior-based approaches, including the dark channel prior (DCP)~\cite{he2010single} and the color attenuation prior (CAP)~\cite{zhu2015fast}; a CNN-based method, FFA-Net~\cite{qin2020ffa}; and a transformer-based method, DehazeFormer~\cite{song2023vision}. These methods are evaluated on the MMAUD dataset~\cite{yuan2024mmaud} to identify the most effective dehazing configuration for the subsequent downstream detection and tracking experiments. Figure~\ref{fig:restoration} presents a qualitative comparison of restored outputs across fog severities, while Table~\ref{tab:restoration_metrics} reports the corresponding quantitative restoration metrics. Based on its superior restoration performance in the present sky-dominant UAV imagery, DehazeFormer was selected to generate the restored datasets used in the downstream detection and tracking experiments.

For a restored image $\hat{\mathbf{I}}$ and its ground truth $\mathbf{I}$ of size $H\times W$, the mean squared error is
\begin{equation}
\mathrm{MSE}(\mathbf{I},\hat{\mathbf{I}})=
\frac{1}{HW}\sum_{i=1}^{H}\sum_{j=1}^{W}\left(\mathbf{I}_{ij}-\hat{\mathbf{I}}_{ij}\right)^2.
\end{equation}
The peak signal-to-noise ratio (PSNR) is then
\begin{equation}
\mathrm{PSNR}(\mathbf{I},\hat{\mathbf{I}})=
10\log_{10}\left(\frac{L^2}{\mathrm{MSE}(\mathbf{I},\hat{\mathbf{I}})}\right),
\end{equation}
where $L$ is the maximum possible pixel value.

The structural similarity index (SSIM) compares luminance, contrast, and structure between local windows $\mathbf{x}$ and $\mathbf{y}$:
\begin{equation}
\mathrm{SSIM}(\mathbf{x},\mathbf{y})=
\frac{(2\mu_x\mu_y+C_1)(2\sigma_{xy}+C_2)}
{(\mu_x^2+\mu_y^2+C_1)(\sigma_x^2+\sigma_y^2+C_2)},
\end{equation}
where $\mu_x$ and $\mu_y$ are local means, $\sigma_x^2$ and $\sigma_y^2$ are local variances, and $\sigma_{xy}$ is the local covariance~\cite{wang2004image}. Higher PSNR and SSIM indicate better restoration quality.

\begin{table}[H]
\centering
\caption{Quantitative restoration performance of CAP, DCP, FFANet, and DehazeFormer under different scattering coefficients $\beta$ on the MMAUD dataset. The best result in each row is shown in bold.}
\label{tab:restoration_metrics}
\small
\begin{tabular}{c cc cc cc cc}
\toprule
\multirow{2}{*}{$\beta$} 
& \multicolumn{2}{c}{CAP} 
& \multicolumn{2}{c}{DCP} 
& \multicolumn{2}{c}{FFANet} 
& \multicolumn{2}{c}{DehazeFormer} \\
\cmidrule(lr){2-3} \cmidrule(lr){4-5} \cmidrule(lr){6-7} \cmidrule(lr){8-9}
& PSNR & SSIM & PSNR & SSIM & PSNR & SSIM & PSNR & SSIM \\
\midrule
0.4 & 17.4 & 0.756 & 12.6 & 0.641 & 33.9 & 0.983 & \textbf{39.3} & \textbf{0.996} \\
0.8 & 21.3 & 0.814 & 13.7 & 0.715 & 28.2 & 0.978 & \textbf{34.1} & \textbf{0.992} \\
1.6 & 19.1 & 0.870 & 16.9 & 0.833 & 26.8 & 0.965 & \textbf{29.9} & \textbf{0.983} \\
2.4 & 14.7 & 0.837 & 22.3 & 0.890 & 24.9 & 0.952 & \textbf{27.3} & \textbf{0.972} \\
3.6 & 11.7 & 0.767 & 16.2 & 0.858 & 23.2 & 0.930 & \textbf{23.9} & \textbf{0.952} \\
\bottomrule
\end{tabular}
\end{table}

\subsection{Detection and Training Regimes}

Detection experiments were conducted using the Ultralytics implementation of the YOLO11 family~\cite{yolo11_ultralytics}, including YOLO11n, YOLO11s, and YOLO11m. All detectors were trained and evaluated on the CfAR dataset~\cite{pereira2024infrared}, which contains 5,834 images. The same 80\%/10\%/10\% split was used throughout the study for training, validation, and testing. All training experiments were conducted using an NVIDIA RTX A2000 GPU. For each detector variant, five training configurations were considered, namely Clean, 30\% Fog, 50\% Fog, 70\% Fog, and 100\% Fog. Each trained detector was evaluated on 11 test sets, including one clean test set, five foggy test sets corresponding to $\beta \in \{0.4, 0.8, 1.6, 2.4, 3.6\}$, and five restored test sets generated by applying the selected DehazeFormer model to the foggy images. The main training settings are summarized in Table~\ref{tab:yolo_train_config}. Unless otherwise stated, the remaining parameters followed the default Ultralytics training configuration.

The experiments were designed to study both clean-trained detection and fog-inclusive training within a unified framework. In the clean-trained setting, each detector was trained only on clean images and then evaluated on the clean test set, five fog-degraded test sets, and the five corresponding restored test sets. This setting isolates the effect of test-time restoration when the detector has not been exposed to fog during training.

In the fog-inclusive setting, a specified proportion of the clean training images was randomly replaced by fog-degraded versions of the same images. For each replaced sample, one of the five fog-severity levels was selected at random, yielding an approximately uniform distribution of fog densities within the replaced subset. This design makes it possible to examine how increasing fog exposure during training affects robustness at inference time.

\begin{table}[H]
\centering
\caption{Main training settings used for the YOLO11 detection experiments.}
\label{tab:yolo_train_config}
\small
\begin{tabular}{ll}
\toprule
\textbf{Setting} & \textbf{Value} \\
\midrule
Detector variants & YOLO11n, YOLO11s, YOLO11m \\
Pretrained initialization & Yes \\
Input image size & 640 \\
Epochs & 100 \\
Batch size & 4 \\
Optimizer & AdamW \\
Learning rate & 0.002 \\
Momentum & 0.9 \\
Weight decay & 0.0005 \\
Automatic mixed precision & Enabled \\
Main augmentations & Mosaic, horizontal flip, HSV jitter, translation, scaling, \\
& random erasing, and RandAugment \\
\bottomrule
\end{tabular}
\end{table}

\subsection{Tracking-by-Detection}

Tracking was performed in a tracking-by-detection setting using ByteTrack~\cite{zhang2022bytetrack} and BoT-SORT~\cite{aharon2022bot}, which associate frame-level detection boxes over time to form target trajectories. The trackers operated on detections produced by the evaluated YOLO11n, YOLO11s, and YOLO11m detectors. To keep the tracking study focused while still reflecting the main detection trends, three representative detector training configurations were selected for tracking, namely Clean, 50\% Fog, and 100\% Fog.

Tracking was evaluated on video sequences from the DUT Anti-UAV dataset~\cite{zhao2022vision}. In addition to the original clean sequences, corresponding fog-degraded and restored versions were generated using the proposed fog synthesis and dehazing pipeline. This design allows the effect of adverse visibility and restoration to be examined in the temporal setting while preserving the same scene content and target motion.

\section{Results}
The results are organized into three parts; (i) the impact of synthetic fog and subsequent image dehazing on detection performance, (ii) the effect of fog-augmented detector training on detection robustness, and (iii) an evaluation of tracking-by-detection performance using both clean-trained and fog-augmented detectors. Note that, for all detection and tracking experiments, the fog-degraded and restored images (and videos) are generated from the same underlying clean images (and videos), ensuring that comparisons are made on identical scenes.

\subsection{Effect of Synthetic Fog and Image Dehazing on Detection Performance}
\begin{figure}[!htbp]
\includegraphics[width=\textwidth]{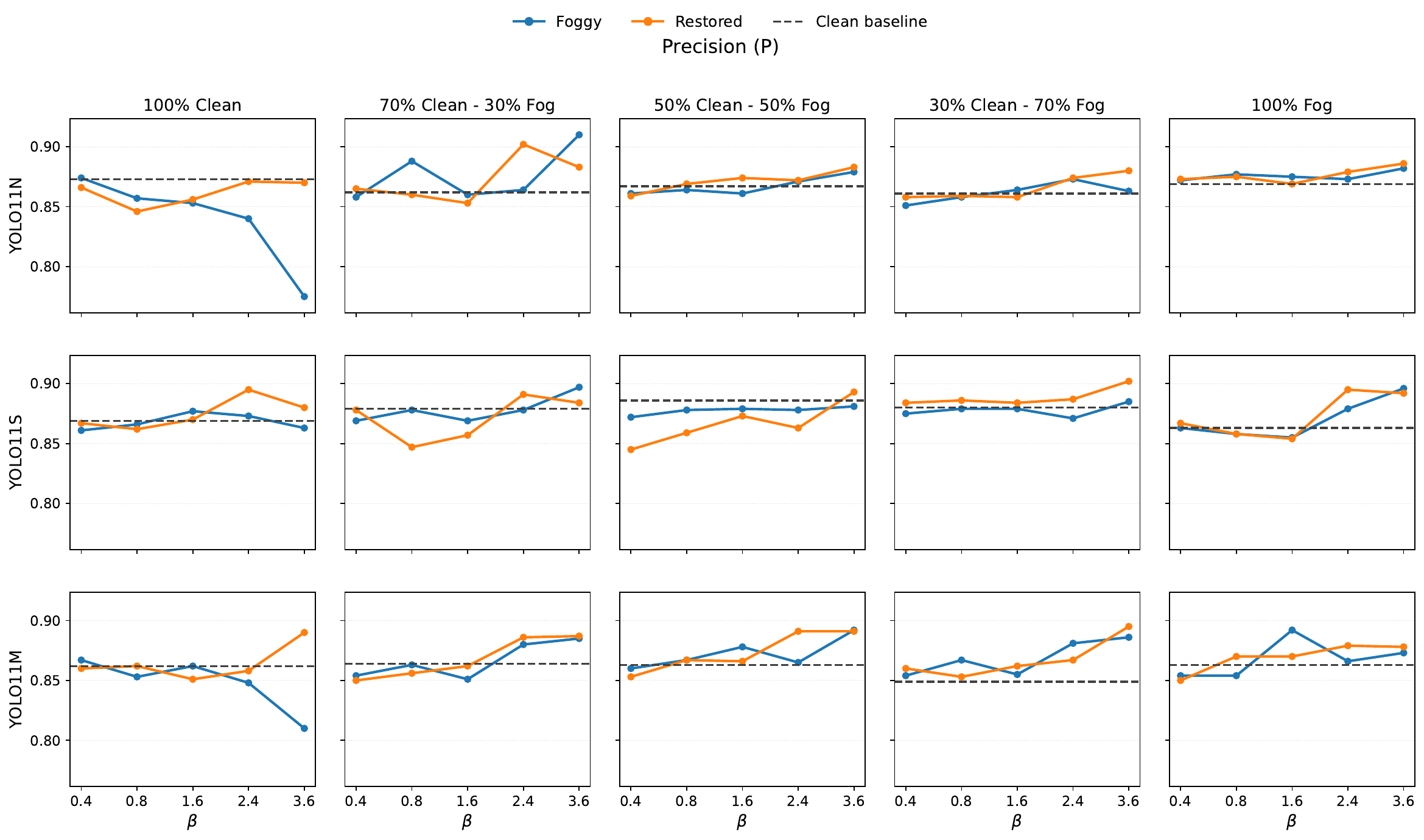}
\caption{Precision (\textit{P}) of YOLO11n, YOLO11s, and YOLO11m under different training-set compositions and evaluation conditions. Rows correspond to detector variants, columns correspond to training configurations, while the curves show performance on foggy and restored test sets across increasing fog severity. The dashed line denotes performance on the clean test set.}
\label{fig:P}
\end{figure}

\begin{figure}[!htbp]
\includegraphics[width=\textwidth]{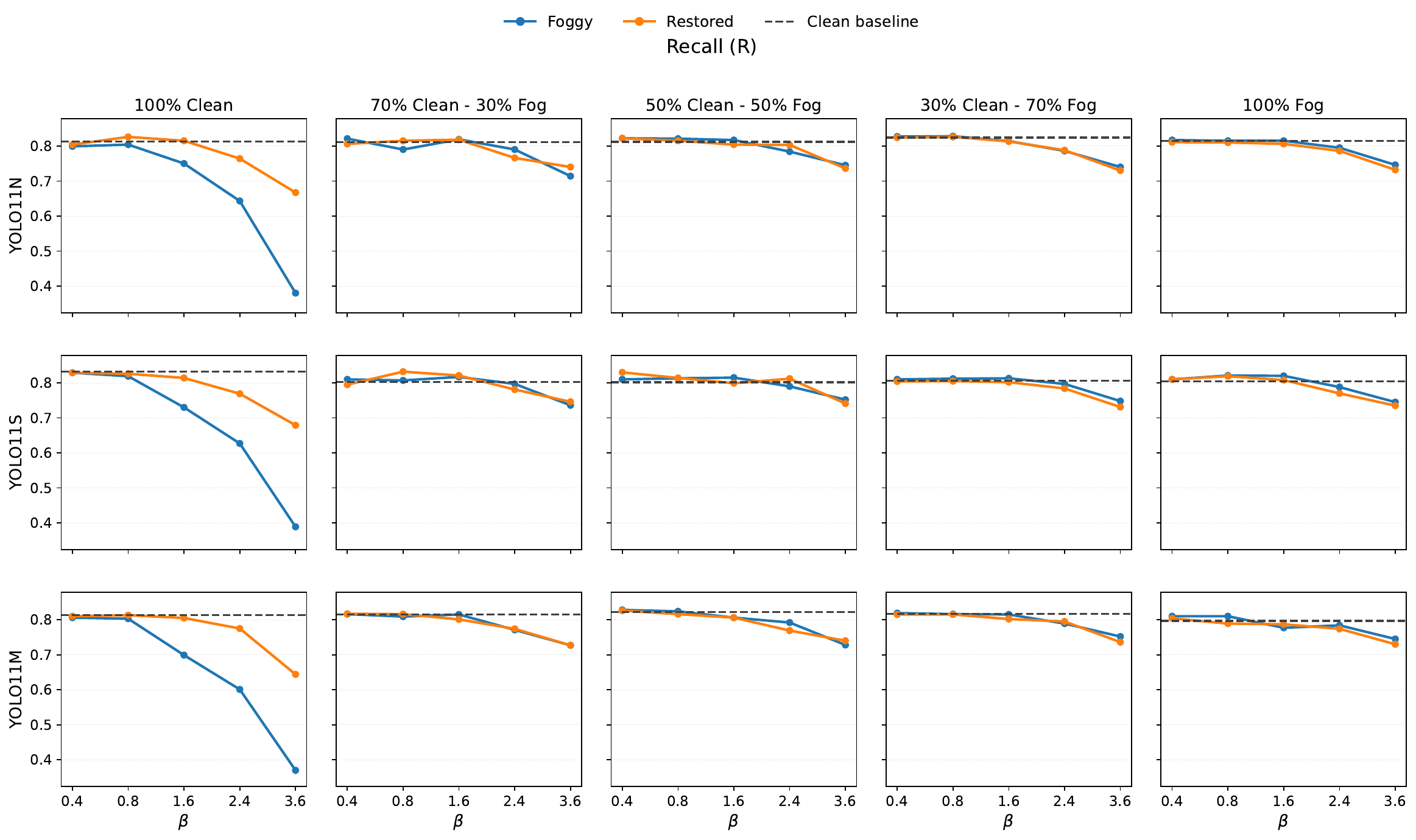}
\caption{Recall (\textit{R}) of the evaluated YOLO11 variants, under the same training and evaluation setup as in Figure~\ref{fig:P}.}
\label{fig:R}
\end{figure}

\begin{figure}[!htbp]
\includegraphics[width=\textwidth]{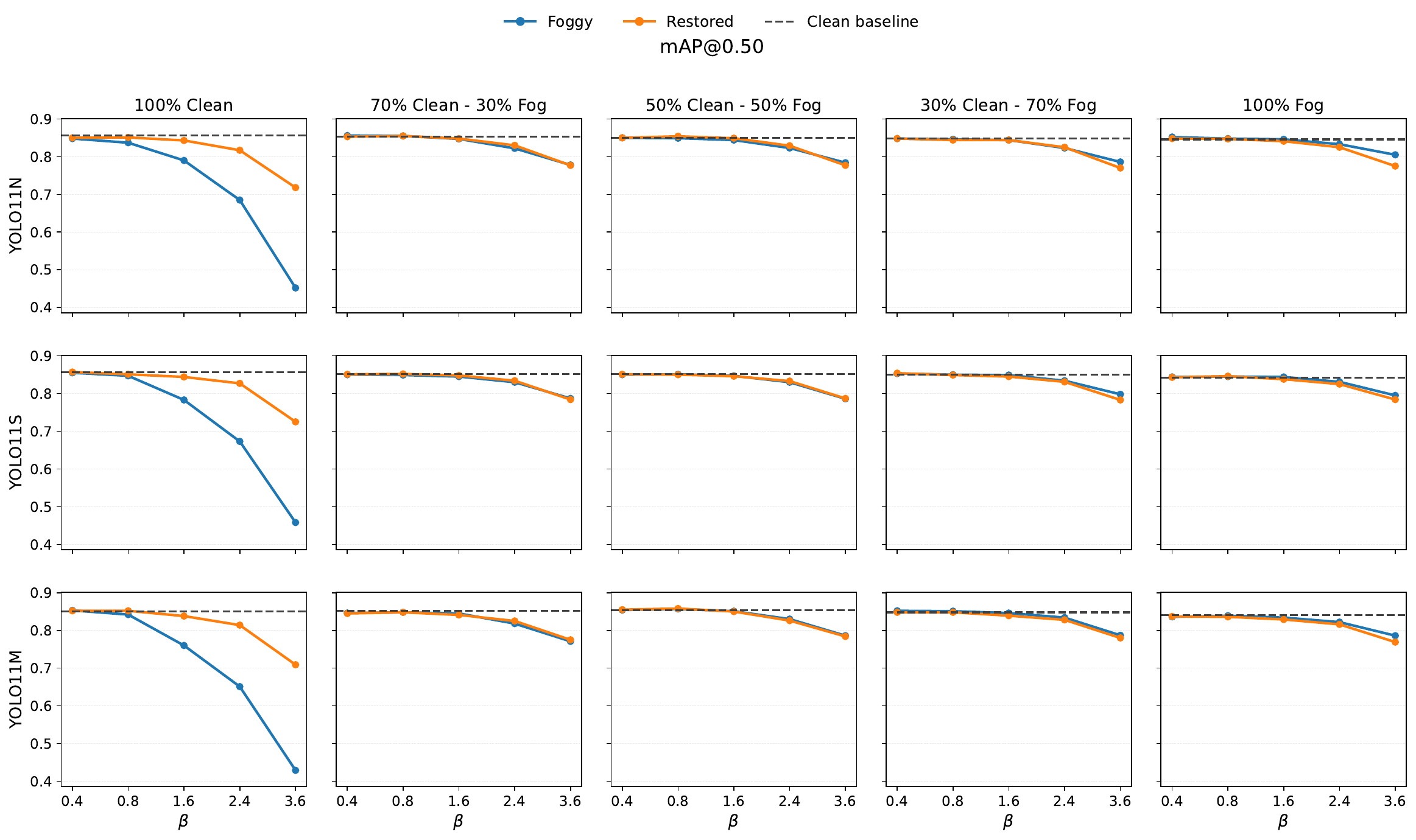}
\caption{Detection performance in terms of mAP@0.50 for the YOLO11 variants, under the same training and evaluation setup as in Figure~\ref{fig:P}.}
\label{fig:mAP50}
\end{figure}

\begin{figure}[!htbp]
\includegraphics[width=\textwidth]{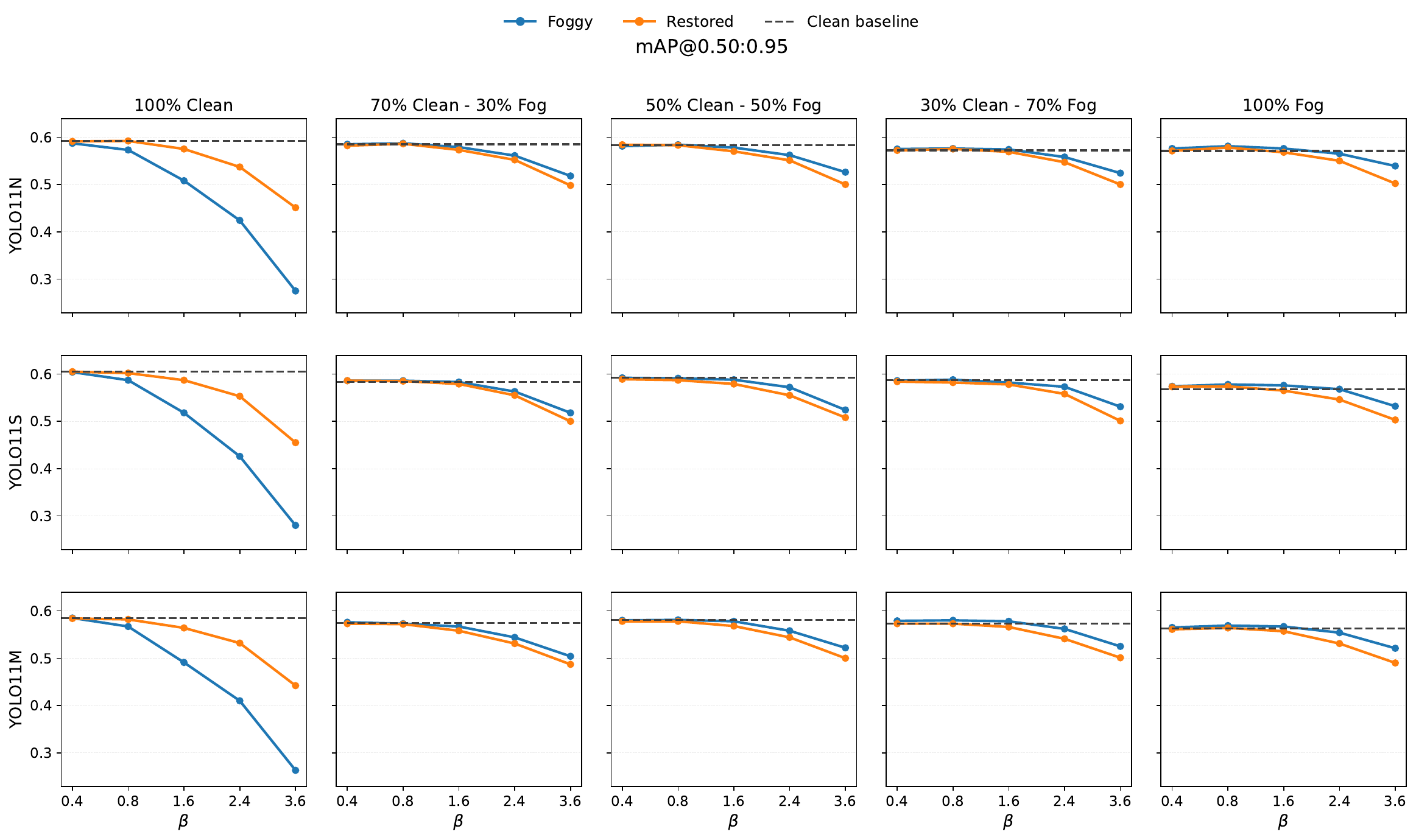}
\caption{Detection performance in terms of mAP@0.50:0.95 for the YOLO11 variants, under the same training and evaluation setup as in Figure~\ref{fig:P}.}
\label{fig:mAP50-95}
\end{figure}

To isolate the contribution of test-time image restoration, the analysis first considers the detector trained exclusively on clean images, corresponding to the first column in Figures~\ref{fig:P}--\ref{fig:mAP50-95}. In this setting, the detector is exposed during training only to the clean imaging domain, and any degradation observed on foggy test images can therefore be attributed primarily to domain shift induced by visibility loss, contrast attenuation, and partial suppression of fine target structure. The resulting trends show that increasing fog severity leads to a pronounced decline in detection performance on foggy inputs, whereas the restored images consistently recover part of the lost performance, particularly at moderate and severe fog levels.

Let TP, FP, and FN denote the numbers of true positives, false positives, and false negatives, respectively. Precision and recall are defined as
\begin{equation}
\mathrm{Precision} = \frac{\mathrm{TP}}{\mathrm{TP}+\mathrm{FP}},
\end{equation}
\begin{equation}
\mathrm{Recall} = \frac{\mathrm{TP}}{\mathrm{TP}+\mathrm{FN}}.
\end{equation}
Average precision (AP) is the area under the precision--recall curve,
\begin{equation}
\mathrm{AP} = \int_{0}^{1} p(r)\,dr,
\end{equation}
where $p(r)$ denotes precision as a function of recall~\cite{everingham2010pascal}. For the present single-class detector, the mean average precision at IoU threshold 0.50 is reported as mAP@0.50, while mAP@0.50:0.95 denotes the mean average precision averaged over IoU thresholds from 0.50 to 0.95 in increments of 0.05.

The clean-trained results reveal two distinct behaviors. First, recall drops much more sharply than precision as fog severity increases, especially for the heaviest degradation levels. This indicates that the dominant failure mode under fog is not a substantial rise in spurious detections, but rather an increase in missed targets, i.e., false negatives. This interpretation is consistent with the visual character of the data where dense fog reduces target--background contrast and suppresses discriminative local structure, making the UAV progressively harder to localize and recognize. Since recall is directly penalized by false negatives, it deteriorates rapidly once the target becomes weak or partially indistinguishable from the background. In contrast, precision remains comparatively more stable because the detector still produces relatively few high-confidence false alarms, even when its ability to recover the true target is impaired. The fluctuations observed in precision are therefore expected. When the number of detections is small, modest changes in the balance between true positives and false positives can noticeably affect precision, even if the dominant degradation mechanism is still missed detection.  

Second, mAP@0.50 and mAP@0.50:0.95 closely follow the recall trend. This is also expected, since a systematic increase in false negatives reduces the attainable recall range and compresses the precision--recall curve, thereby lowering AP and its mean variants. The stronger drop in mAP@0.50:0.95 compared with mAP@0.50 under severe fog further suggests that fog affects not only target detectability but also localization quality, because the stricter IoU thresholds used in mAP@0.50:0.95 are more sensitive to degraded box alignment. In all three YOLO11 variants, restored images remain consistently above the corresponding foggy images in the clean-trained column, showing that restoration partially reduces the clean-to-fog domain gap and recovers target cues that are otherwise lost in the degraded inputs. However, the restored curves still remain below the clean baseline, indicating that restoration is not able to fully reconstruct the original appearance distribution seen during training.

\begin{figure}[H]
\includegraphics[width=\textwidth,trim={0.7cm 0.4cm 0.7cm 0.4cm},clip]{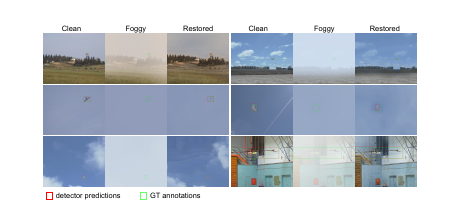}
\caption{Representative qualitative examples showing cases in which the YOLO11n detector trained on the clean training and validation sets fails to detect the target in fog-degraded images but successfully detects it again after DehazeFormer-based restoration. The columns correspond to Clean images, Foggy images with $\beta = 3.6$, and Restored images. Red and green bounding boxes indicate detector predictions and ground-truth annotations, respectively.}
\label{fig:triptych}
\end{figure}

Figure~\ref{fig:triptych} provides qualitative support for this interpretation. The examples show cases in which the clean-trained detector fails on heavily fogged inputs but succeeds again after restoration, demonstrating that restoration can recover sufficient visual evidence for the detector to re-establish the target response. At the same time, such recovery is not uniform across all images or all severities, which is consistent with the metric plots and indicates that improved perceptual appearance does not guarantee proportional downstream gains. Restoration can reduce false negatives by improving target visibility, but it may also alter local image statistics and object boundaries in a way that prevents perfect recovery of clean-domain detector behavior.

\subsection{Effect of Fog-Augmented Training}\label{subsec:Fog-Augmented}
The remaining four columns in Figures~\ref{fig:P}--\ref{fig:mAP50-95} examine the effect of incorporating foggy samples into training. Compared with the clean-trained case, the most prominent outcome is that the foggy and restored curves become substantially closer to the clean baseline and also closer to each other as the fog proportion in training increases. This indicates that fog-inclusive training is the primary mechanism for improving robustness under adverse visibility, because it directly exposes the detector to the appearance distortions induced by fog and reduces the train--test mismatch encountered at inference time. Across all three YOLO11 variants, the severe collapse observed for recall and both mAP metrics in the clean-trained setting is greatly suppressed once foggy images are included during training, especially for the 50\%, 70\%, and 100\% fog configurations. 

The behavior of recall is particularly informative. Under clean-only training, recall decreases sharply with fog severity, reflecting a rapid growth in false negatives. Under mixed training, however, recall degrades much more gradually, which implies that the detector is less likely to miss the UAV even when target contrast and structure are degraded by fog. This improvement propagates directly to mAP@0.50 and mAP@0.50:0.95, whose trajectories largely mirror the recall curves across models and training compositions. Such similarity between recall and mAP is expected in this regime because the dominant change is whether the UAV is detected at all, rather than a dramatic rise in false positive detections. Once fog-aware training reduces missed detections, the precision--recall curve remains closer to the clean reference, and both mAP measures remain correspondingly more stable. The remaining gap between mAP@0.50 and mAP@0.50:0.95 at high fog levels suggests that localization quality is still affected under severe degradation, even when target presence is recovered more reliably. 

Precision behaves differently from recall and mAP. Across most training settings, it remains relatively high and exhibits mild fluctuations rather than a monotonic decline, which is consistent with a regime in which fog affects detection primarily through missed targets rather than through a large increase in false positives. Since precision depends on the ratio $\mathrm{TP}/(\mathrm{TP}+\mathrm{FP})$, small variations in confidence ranking, duplicated boxes, or a limited number of additional false positives can produce visible oscillations even when the overall degradation trend is smooth. This also helps explain why the 30\% fog-trained models show more fluctuation than the 100\% fog-trained models. With less fog exposure during training, the detector remains more sensitive to severity-dependent appearance changes, whereas full fog training reduces the domain mismatch and leads to more stable precision across fog levels. For this reason, precision is less visually aligned with fog severity than recall and mAP and should be interpreted together with those metrics rather than in isolation.

An important observation is that restoration is clearly more beneficial in the clean-trained analysis than in the mixed-training analysis. In the first column, restored inputs are consistently above foggy inputs because dehazing partially moves the degraded images back toward the clean training domain. In the mixed-training columns, however, foggy and restored performance often become very close, and in some cases foggy performance is slightly better than restored performance. This is also reasonable; once the detector has learned directly from foggy data, raw foggy images are no longer strongly out-of-distribution, and restoration is not guaranteed to provide an additional advantage. On the contrary, restoration may alter object boundaries, local contrast, or fine-scale textures in a way that is not perfectly aligned with either the clean or foggy training distributions. As a result, the benefit of restoration becomes conditional rather than universal. The figures therefore support a nuanced conclusion that restoration is most useful when the detector has not been trained to handle fog, whereas fog-inclusive training reduces the dependence on restoration by learning robustness directly from the degraded domain. 

Comparing across detector sizes, all three YOLO11 variants exhibit the same qualitative behavior, indicating that the conclusions are not specific to a single model scale. The larger models generally maintain slightly stronger adverse-weather performance, but the dominant factor in the present study is not architecture size alone; rather, it is the interaction between training-set composition and test-domain shift. The clean-trained models suffer the greatest mismatch and therefore benefit most from restoration, whereas the mixed-trained models are more resilient to fog and consequently show smaller foggy--restored gaps. Taken together, these results indicate that fog-aware training is the more reliable route to robustness, while restoration acts as an auxiliary mechanism whose effectiveness depends on how strongly the detector has already adapted to fog during training. 

To quantify robustness across fog severities, the coefficient of variation of mAP is defined as
\begin{equation}
\mathrm{CV}_{\mathrm{mAP}}=\frac{\sigma_{\mathrm{mAP}}}{\mu_{\mathrm{mAP}}},
\label{eq:cv_map}
\end{equation}
where $\mu_{\mathrm{mAP}}$ and $\sigma_{\mathrm{mAP}}$ denote the mean and standard deviation of mAP over the five fog severities. Lower $\mathrm{CV}_{\mathrm{mAP}}$ indicates greater stability under increasing degradation and complements the mean mAP by measuring sensitivity to severity changes.

\begin{figure}[H]
\includegraphics[width=\textwidth]{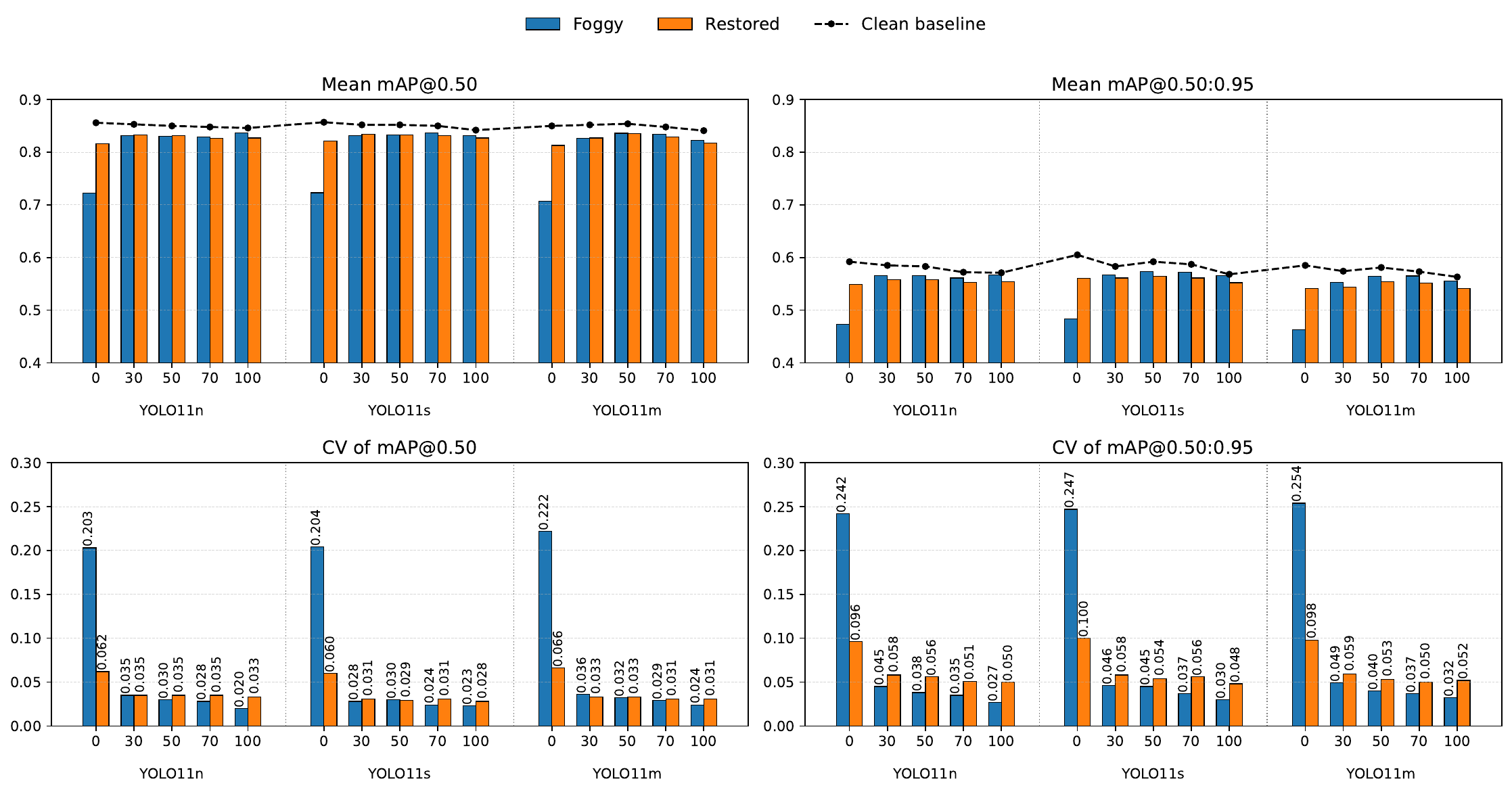}
\caption{Detection performance and stability of YOLO11n, YOLO11s, and YOLO11m trained with different percentages of foggy images. The top row shows the mean mAP@0.50 and mAP@0.50:0.95 over the five foggy and five restored test sets, while the dashed line denotes the clean-test baseline. The bottom row shows the corresponding coefficient of variation (CV) across fog severities.}
\label{fig:cv_map_models}
\end{figure}

Figure~\ref{fig:cv_map_models} compares the mean performance and stability of YOLO11n, YOLO11s, and YOLO11m trained with different proportions of foggy images. For all three detector variants, the clean-trained models yield the lowest mean performance on the foggy test sets in both mAP@0.50 and mAP@0.50:0.95, together with the highest CV values. This shows that training only on clean images leads not only to lower adverse-weather performance but also to substantially greater sensitivity to fog severity. Test-time restoration improves the clean-trained detectors, increasing the mean mAP and reducing the corresponding CV relative to the foggy inputs, although the restored means remain below the clean-test baseline.

In contrast, incorporating foggy images into training consistently improves robustness. Across YOLO11n, YOLO11s, and YOLO11m, fog-inclusive training brings the foggy and restored means closer to the clean baseline and substantially reduces the CV values relative to the clean-trained case. The largest gain therefore comes from the transition from clean-only to fog-aware training, while the differences among the mixed and fully foggy-trained models are comparatively smaller. Moreover, for the fog-inclusive models, the foggy and restored CV values become similar, indicating that once the detector has adapted to fog during training, restoration provides a smaller and less consistent stability benefit.

Overall, the three detector variants exhibit the same qualitative trend. Clean-only training yields the weakest and least stable performance under fog, whereas fog-inclusive training improves both the average mAP and its stability across severities. These results indicate that, for downstream tracking-by-detection, detector selection should be based not only on mean performance but also on stability, since lower CV implies more reliable behavior under changing visibility conditions.

\subsection{Tracking Performance under Synthetic Fog and Image Restoration}
\label{subsec:tracking_results}

To assess how detector robustness transfers to tracking-by-detection, ByteTrack and BoT-SORT are evaluated on DUT Anti-UAV videos using YOLO11n, YOLO11s, and YOLO11m trained under three representative settings, namely clean, 50\% foggy, and 100\% foggy. Tracking is then evaluated on clean, foggy, and restored versions of video sequences from the DUT Anti-UAV dataset. Tracking results are summarized in Tables~\ref{tab:dut_tracking_summary_bytetrack} and~\ref{tab:dut_tracking_summary_botsort}.

Multiple object tracking accuracy (MOTA) is defined as
\begin{equation}
\mathrm{MOTA} = 1 - \frac{\sum_t \left(\mathrm{FN}_t+\mathrm{FP}_t+\mathrm{IDSW}_t\right)}
{\sum_t \mathrm{GT}_t},
\end{equation}
where $\mathrm{FN}_t$, $\mathrm{FP}_t$, and $\mathrm{IDSW}_t$ denote false negatives, false positives, and identity switches at time $t$, and $\mathrm{GT}_t$ is the number of ground-truth objects~\cite{bernardin2008evaluating}. Identity-preserving tracking quality is measured by IDF1,
\begin{equation}
\mathrm{IDF1}=
\frac{2\,\mathrm{IDTP}}{2\,\mathrm{IDTP}+\mathrm{IDFP}+\mathrm{IDFN}},
\end{equation}
where IDTP, IDFP, and IDFN denote identity true positives, identity false positives, and identity false negatives, respectively~\cite{ristani2016performance}. Higher MOTA and IDF1 indicate better tracking performance.

Across both trackers, the dominant factor is the robustness of the underlying detector. Sequences that remain detectable under clean and light-fog conditions generally preserve high MOTA and IDF1, whereas performance drops sharply once fog causes missed detections. This is consistent with the detection results, which show that adverse weather affects tracking primarily through increased false negatives, which the tracker cannot recover once the target response is lost.

Clean-trained detectors are consistently the most vulnerable under severe fog. This is particularly evident in Video~02 and Video~10, where several clean-trained models show severe degradation, including near-zero or negative MOTA in the highest-fog condition. In contrast, fog-inclusive training, especially the 50\% and 100\% fog configurations, generally preserves stronger tracking performance across all three YOLO11 variants. The main gain therefore appears to come from improved detector robustness rather than from the tracker alone.

The effect of restoration is beneficial but not uniform. For clean-trained detectors, restored videos often recover performance relative to the corresponding foggy videos, particularly at moderate and severe fog levels. After fog-aware training, however, the gap between foggy and restored tracking usually becomes smaller, indicating that restoration provides less additional benefit once the detector has already adapted to degraded inputs. In some cases, restored results remain close to or slightly below the foggy results, showing that improved visual appearance does not always translate into stronger tracking.

\begin{table}[H]
\centering
\caption{ByteTrack tracking performance on DUT Anti-UAV sample videos under clean, fog-degraded, and restored conditions.}
\label{tab:dut_tracking_summary_bytetrack}
\scriptsize
\setlength{\tabcolsep}{2.2pt}
\renewcommand{\arraystretch}{1.10}
\resizebox{\textwidth}{!}{%
\begin{tabular}{c c cc cccc cccc cccc}
\toprule
\multirow{3}{*}{\textbf{Video}} &
\multirow{3}{*}{\textbf{\shortstack{YOLO\\Trained\\model}}} &
\multicolumn{2}{c}{\textbf{Clean}} &
\multicolumn{4}{c}{\textbf{$\beta$ = 0.4}} &
\multicolumn{4}{c}{\textbf{$\beta$ = 1.6}} &
\multicolumn{4}{c}{\textbf{$\beta$ = 3.6}} \\
\cmidrule(lr){3-4}
\cmidrule(lr){5-8}
\cmidrule(lr){9-12}
\cmidrule(lr){13-16}
&
& \multicolumn{2}{c}{ }
& \multicolumn{2}{c}{Foggy} & \multicolumn{2}{c}{Restored}
& \multicolumn{2}{c}{Foggy} & \multicolumn{2}{c}{Restored}
& \multicolumn{2}{c}{Foggy} & \multicolumn{2}{c}{Restored} \\
\cmidrule(lr){5-6}\cmidrule(lr){7-8}
\cmidrule(lr){9-10}\cmidrule(lr){11-12}
\cmidrule(lr){13-14}\cmidrule(lr){15-16}
&
& MOTA & IDF1
& MOTA & IDF1 & MOTA & IDF1
& MOTA & IDF1 & MOTA & IDF1
& MOTA & IDF1 & MOTA & IDF1 \\
\midrule

\multirow{9}{*}{02} & 11n clean
& 1.000 & 1.000
& 0.976 & 0.988 & 0.988 & 0.994
& 0.265 & 0.419 & 0.988 & 0.994
& 0.000 & 0.000 & 0.277 & 0.434 \\
& 11n 50\% foggy
& 1.000 & 1.000
& 0.988 & 0.994 & 0.988 & 0.994
& 0.988 & 0.994 & 0.988 & 0.994
& 0.566 & 0.687 & 0.663 & 0.797 \\
& 11n 100\% foggy
& 1.000 & 1.000
& 0.988 & 0.994 & 0.988 & 0.994
& 0.988 & 0.994 & 0.976 & 0.988
& 0.578 & 0.733 & 0.494 & 0.661 \\
& 11s clean
& 1.000 & 1.000
& 0.988 & 0.994 & 0.988 & 0.994
& 0.988 & 0.994 & 0.976 & 0.988
& 0.000 & 0.000 & 0.639 & 0.786 \\
& 11s 50\% foggy
& 1.000 & 1.000
& 0.988 & 0.994 & 0.988 & 0.994
& 0.988 & 0.994 & 0.988 & 0.994
& 0.578 & 0.707 & 0.108 & 0.479 \\
& 11s 100\% foggy
& 1.000 & 1.000
& 0.988 & 0.994 & 0.988 & 0.994
& 0.988 & 0.994 & 0.988 & 0.994
& 0.771 & 0.871 & 0.759 & 0.863 \\
& 11m clean
& 0.892 & 0.945
& 0.964 & 0.982 & 0.880 & 0.939
& 0.843 & 0.915 & 0.928 & 0.963
& 0.000 & 0.000 & -0.578 & 0.166 \\
& 11m 50\% foggy
& 1.000 & 1.000
& 0.988 & 0.994 & 0.988 & 0.994
& 0.759 & 0.863 & 0.988 & 0.994
& 0.373 & 0.544 & 0.422 & 0.593 \\
& 11m 100\% foggy
& 1.000 & 1.000
& 0.988 & 0.994 & 0.976 & 0.988
& 0.843 & 0.919 & 0.988 & 0.994
& 0.181 & 0.306 & 0.711 & 0.831 \\
\midrule

\multirow{9}{*}{03} & 11n clean
& 0.540 & 0.701
& 0.570 & 0.726 & 0.540 & 0.701
& 0.260 & 0.394 & 0.660 & 0.795
& 0.160 & 0.276 & 0.550 & 0.710 \\
& 11n 50\% foggy
& 0.930 & 0.964
& 0.910 & 0.953 & 0.930 & 0.964
& 0.810 & 0.895 & 0.910 & 0.953
& 0.260 & 0.413 & 0.710 & 0.830 \\
& 11n 100\% foggy
& 0.520 & 0.684
& 0.500 & 0.667 & 0.540 & 0.701
& 0.520 & 0.588 & 0.630 & 0.773
& 0.280 & 0.438 & 0.430 & 0.458 \\
& 11s clean
& 0.920 & 0.958
& 0.920 & 0.958 & 0.950 & 0.974
& 0.960 & 0.980 & 0.920 & 0.958
& 0.250 & 0.270 & 0.810 & 0.895 \\
& 11s 50\% foggy
& 0.970 & 0.985
& 0.970 & 0.985 & 0.970 & 0.985
& 0.960 & 0.980 & 0.970 & 0.985
& 0.830 & 0.891 & 0.730 & 0.846 \\
& 11s 100\% foggy
& 0.960 & 0.980
& 0.970 & 0.985 & 0.970 & 0.985
& 0.970 & 0.985 & 0.940 & 0.969
& 0.940 & 0.969 & 0.760 & 0.864 \\
& 11m clean
& 0.880 & 0.847
& 0.970 & 0.985 & 0.860 & 0.866
& 0.970 & 0.985 & 0.910 & 0.953
& 0.210 & 0.347 & 0.830 & 0.907 \\
& 11m 50\% foggy
& 0.980 & 0.990
& 0.980 & 0.990 & 0.980 & 0.990
& 0.970 & 0.985 & 0.960 & 0.980
& 0.920 & 0.958 & 0.890 & 0.942 \\
& 11m 100\% foggy
& 0.900 & 0.947
& 0.930 & 0.964 & 0.930 & 0.964
& 0.960 & 0.980 & 0.930 & 0.964
& 0.900 & 0.947 & 0.780 & 0.876 \\
\midrule

\multirow{9}{*}{06} & 11n clean
& 0.980 & 0.990
& 0.985 & 0.992 & 0.995 & 0.997
& 0.840 & 0.913 & 0.995 & 0.997
& 0.245 & 0.394 & 0.380 & 0.551 \\
& 11n 50\% foggy
& 0.995 & 0.997
& 0.995 & 0.997 & 0.995 & 0.997
& 0.925 & 0.943 & 0.995 & 0.997
& 0.725 & 0.841 & 0.525 & 0.510 \\
& 11n 100\% foggy
& 0.975 & 0.987
& 0.955 & 0.977 & 0.965 & 0.982
& 0.910 & 0.953 & 0.915 & 0.956
& 0.765 & 0.867 & 0.465 & 0.565 \\
& 11s clean
& 0.995 & 0.997
& 0.995 & 0.997 & 0.995 & 0.997
& 0.980 & 0.990 & 0.995 & 0.997
& 0.140 & 0.246 & 0.485 & 0.470 \\
& 11s 50\% foggy
& 0.990 & 0.995
& 0.990 & 0.995 & 0.990 & 0.995
& 0.970 & 0.985 & 0.995 & 0.997
& 0.540 & 0.619 & 0.360 & 0.513 \\
& 11s 100\% foggy
& 0.980 & 0.990
& 0.975 & 0.987 & 0.980 & 0.990
& 0.950 & 0.974 & 0.985 & 0.992
& 0.960 & 0.980 & 0.645 & 0.580 \\
& 11m clean
& 0.995 & 0.997
& 0.995 & 0.997 & 0.995 & 0.997
& 0.995 & 0.997 & 0.995 & 0.997
& 0.220 & 0.361 & 0.790 & 0.883 \\
& 11m 50\% foggy
& 0.995 & 0.997
& 0.995 & 0.997 & 0.995 & 0.997
& 0.990 & 0.995 & 0.995 & 0.997
& 0.800 & 0.889 & 0.630 & 0.526 \\
& 11m 100\% foggy
& 0.975 & 0.987
& 0.995 & 0.997 & 0.985 & 0.992
& 0.995 & 0.997 & 0.995 & 0.997
& 0.410 & 0.509 & 0.635 & 0.537 \\
\midrule

\multirow{9}{*}{10} & 11n clean
& 0.245 & 0.200
& 0.299 & 0.194 & 0.302 & 0.205
& 0.107 & 0.095 & 0.246 & 0.182
& 0.000 & 0.000 & 0.003 & 0.005 \\
& 11n 50\% foggy
& 0.264 & 0.191
& 0.275 & 0.190 & 0.268 & 0.192
& 0.244 & 0.174 & 0.252 & 0.184
& 0.052 & 0.059 & 0.053 & 0.051 \\
& 11n 100\% foggy
& 0.321 & 0.210
& 0.336 & 0.342 & 0.332 & 0.341
& 0.325 & 0.209 & 0.312 & 0.208
& 0.078 & 0.062 & 0.050 & 0.066 \\
& 11s clean
& 0.051 & 0.174
& 0.272 & 0.174 & -0.066 & 0.156
& 0.257 & 0.168 & -0.172 & 0.142
& 0.007 & 0.014 & -0.157 & 0.069 \\
& 11s 50\% foggy
& 0.303 & 0.195
& 0.368 & 0.240 & 0.332 & 0.194
& 0.347 & 0.204 & 0.319 & 0.188
& 0.157 & 0.147 & 0.154 & 0.123 \\
& 11s 100\% foggy
& 0.291 & 0.188
& 0.316 & 0.189 & 0.298 & 0.187
& 0.359 & 0.182 & 0.301 & 0.189
& 0.168 & 0.119 & 0.179 & 0.141 \\
& 11m clean
& -0.225 & 0.138
& 0.156 & 0.176 & 0.100 & 0.165
& 0.111 & 0.108 & 0.002 & 0.150
& 0.019 & 0.034 & 0.036 & 0.079 \\
& 11m 50\% foggy
& 0.302 & 0.207
& 0.311 & 0.207 & 0.303 & 0.205
& 0.339 & 0.258 & 0.277 & 0.191
& 0.079 & 0.079 & 0.050 & 0.068 \\
& 11m 100\% foggy
& 0.260 & 0.205
& 0.232 & 0.209 & 0.267 & 0.204
& 0.161 & 0.175 & 0.241 & 0.195
& 0.038 & 0.045 & 0.049 & 0.059 \\
\bottomrule

\end{tabular}%
}
\end{table}

Sequence difficulty also plays a major role. Video~06 is comparatively stable across most detector and tracker settings, whereas Video~10 is consistently the most challenging sequence for both ByteTrack and BoT-SORT. This suggests that the main limitation is the combined difficulty of small-target detection, sequence content, and severe visibility loss, rather than the tracker choice alone.

ByteTrack and BoT-SORT exhibit broadly similar trends across detector variants and weather conditions. Both benefit from fog-aware detector training and both deteriorate when the detector fails under severe fog. Their differences are sequence- and condition-dependent rather than uniformly favoring one tracker, which is consistent with a tracking-by-detection setting in which missed detections dominate the error budget. Overall, the tracking results indicate that detector robustness is the primary driver of adverse-weather tracking performance, while tracker choice has a secondary but measurable effect.

\begin{table}[H]
\centering
\caption{BoT-SORT tracking performance on DUT Anti-UAV sample videos under clean, fog-degraded, and restored conditions.}
\label{tab:dut_tracking_summary_botsort}
\scriptsize
\setlength{\tabcolsep}{2.2pt}
\renewcommand{\arraystretch}{1.10}
\resizebox{\textwidth}{!}{%
\begin{tabular}{c c cc cccc cccc cccc}
\toprule
\multirow{3}{*}{\textbf{Video}} &
\multirow{3}{*}{\textbf{\shortstack{YOLO\\Trained\\model}}} &
\multicolumn{2}{c}{\textbf{Clean}} &
\multicolumn{4}{c}{\textbf{$\beta$ = 0.4}} &
\multicolumn{4}{c}{\textbf{$\beta$ = 1.6}} &
\multicolumn{4}{c}{\textbf{$\beta$ = 3.6}} \\
\cmidrule(lr){3-4}
\cmidrule(lr){5-8}
\cmidrule(lr){9-12}
\cmidrule(lr){13-16}
&
& \multicolumn{2}{c}{ }
& \multicolumn{2}{c}{Foggy} & \multicolumn{2}{c}{Restored}
& \multicolumn{2}{c}{Foggy} & \multicolumn{2}{c}{Restored}
& \multicolumn{2}{c}{Foggy} & \multicolumn{2}{c}{Restored} \\
\cmidrule(lr){5-6}\cmidrule(lr){7-8}
\cmidrule(lr){9-10}\cmidrule(lr){11-12}
\cmidrule(lr){13-14}\cmidrule(lr){15-16}
&
& MOTA & IDF1
& MOTA & IDF1 & MOTA & IDF1
& MOTA & IDF1 & MOTA & IDF1
& MOTA & IDF1 & MOTA & IDF1 \\
\midrule

\multirow{9}{*}{02} & 11n clean
& 1.000 & 1.000
& 0.976 & 0.988 & 0.988 & 0.994
& 0.253 & 0.404 & 0.988 & 0.994
& 0.000 & 0.000 & 0.277 & 0.434 \\
& 11n 50\% foggy
& 1.000 & 1.000
& 0.988 & 0.994 & 0.988 & 0.994
& 0.988 & 0.994 & 0.988 & 0.994
& 0.614 & 0.667 & 0.675 & 0.806 \\
& 11n 100\% foggy
& 1.000 & 1.000
& 0.988 & 0.994 & 0.988 & 0.994
& 0.988 & 0.994 & 0.976 & 0.988
& 0.578 & 0.733 & 0.494 & 0.661 \\
& 11s clean
& 1.000 & 1.000
& 0.988 & 0.994 & 0.988 & 0.994
& 0.904 & 0.949 & 0.976 & 0.988
& 0.000 & 0.000 & 0.639 & 0.786 \\
& 11s 50\% foggy
& 1.000 & 1.000
& 0.988 & 0.994 & 0.988 & 0.994
& 0.988 & 0.994 & 0.988 & 0.994
& 0.602 & 0.752 & 0.108 & 0.479 \\
& 11s 100\% foggy
& 1.000 & 1.000
& 0.988 & 0.994 & 0.988 & 0.994
& 0.976 & 0.988 & 0.988 & 0.994
& 0.759 & 0.863 & 0.759 & 0.863 \\
& 11m clean
& 0.892 & 0.945
& 0.964 & 0.982 & 0.880 & 0.939
& 0.843 & 0.915 & 0.928 & 0.963
& 0.000 & 0.000 & -0.578 & 0.166 \\
& 11m 50\% foggy
& 1.000 & 1.000
& 0.988 & 0.994 & 0.988 & 0.994
& 0.759 & 0.863 & 0.988 & 0.994
& 0.289 & 0.449 & 0.410 & 0.581 \\
& 11m 100\% foggy
& 1.000 & 1.000
& 0.988 & 0.994 & 0.976 & 0.988
& 0.843 & 0.919 & 0.988 & 0.994
& 0.181 & 0.306 & 0.687 & 0.809 \\
\midrule

\multirow{9}{*}{03} & 11n clean
& 0.530 & 0.693
& 0.660 & 0.795 & 0.570 & 0.726
& 0.270 & 0.406 & 0.520 & 0.484
& 0.050 & 0.074 & 0.210 & 0.288 \\
& 11n 50\% foggy
& 0.930 & 0.964
& 0.920 & 0.958 & 0.940 & 0.969
& 0.640 & 0.398 & 0.770 & 0.413
& 0.180 & 0.230 & 0.230 & 0.281 \\
& 11n 100\% foggy
& 0.570 & 0.726
& 0.550 & 0.710 & 0.560 & 0.718
& 0.420 & 0.434 & 0.480 & 0.483
& 0.190 & 0.244 & 0.190 & 0.293 \\
& 11s clean
& 0.930 & 0.964
& 0.920 & 0.958 & 0.950 & 0.974
& 0.890 & 0.398 & 0.820 & 0.396
& 0.110 & 0.122 & 0.280 & 0.259 \\
& 11s 50\% foggy
& 0.970 & 0.985
& 0.970 & 0.985 & 0.970 & 0.985
& 0.900 & 0.394 & 0.860 & 0.387
& 0.770 & 0.807 & 0.280 & 0.263 \\
& 11s 100\% foggy
& 0.970 & 0.985
& 0.970 & 0.985 & 0.970 & 0.985
& 0.910 & 0.402 & 0.820 & 0.396
& 0.860 & 0.899 & 0.300 & 0.261 \\
& 11m clean
& 0.910 & 0.953
& 0.970 & 0.985 & 0.910 & 0.953
& 0.910 & 0.402 & 0.830 & 0.394
& 0.140 & 0.169 & 0.310 & 0.252 \\
& 11m 50\% foggy
& 0.980 & 0.990
& 0.980 & 0.990 & 0.980 & 0.990
& 0.910 & 0.402 & 0.850 & 0.389
& 0.840 & 0.851 & 0.300 & 0.252 \\
& 11m 100\% foggy
& 0.900 & 0.947
& 0.930 & 0.964 & 0.930 & 0.964
& 0.900 & 0.394 & 0.820 & 0.396
& 0.840 & 0.851 & 0.300 & 0.257 \\
\midrule

\multirow{9}{*}{06} & 11n clean
& 0.980 & 0.990
& 0.985 & 0.992 & 0.995 & 0.997
& 0.840 & 0.913 & 0.995 & 0.997
& 0.245 & 0.394 & 0.380 & 0.551 \\
& 11n 50\% foggy
& 0.995 & 0.997
& 0.995 & 0.997 & 0.995 & 0.997
& 0.925 & 0.943 & 0.995 & 0.997
& 0.725 & 0.841 & 0.525 & 0.510 \\
& 11n 100\% foggy
& 0.975 & 0.987
& 0.955 & 0.977 & 0.965 & 0.982
& 0.910 & 0.953 & 0.915 & 0.956
& 0.770 & 0.870 & 0.465 & 0.565 \\
& 11s clean
& 0.995 & 0.997
& 0.995 & 0.997 & 0.995 & 0.997
& 0.980 & 0.990 & 0.995 & 0.997
& 0.140 & 0.246 & 0.485 & 0.470 \\
& 11s 50\% foggy
& 0.990 & 0.995
& 0.990 & 0.995 & 0.990 & 0.995
& 0.970 & 0.985 & 0.995 & 0.997
& 0.535 & 0.615 & 0.360 & 0.513 \\
& 11s 100\% foggy
& 0.980 & 0.990
& 0.975 & 0.987 & 0.980 & 0.990
& 0.950 & 0.974 & 0.985 & 0.992
& 0.960 & 0.980 & 0.645 & 0.580 \\
& 11m clean
& 0.995 & 0.997
& 0.995 & 0.997 & 0.995 & 0.997
& 0.995 & 0.997 & 0.995 & 0.997
& 0.220 & 0.361 & 0.790 & 0.883 \\
& 11m 50\% foggy
& 0.995 & 0.997
& 0.995 & 0.997 & 0.995 & 0.997
& 0.990 & 0.995 & 0.995 & 0.997
& 0.795 & 0.886 & 0.625 & 0.528 \\
& 11m 100\% foggy
& 0.975 & 0.987
& 0.995 & 0.997 & 0.985 & 0.992
& 0.995 & 0.997 & 0.995 & 0.997
& 0.410 & 0.509 & 0.635 & 0.537 \\
\midrule

\multirow{9}{*}{10} & 11n clean
& 0.246 & 0.200
& 0.300 & 0.194 & 0.302 & 0.205
& 0.107 & 0.094 & 0.248 & 0.182
& 0.000 & 0.000 & 0.003 & 0.005 \\
& 11n 50\% foggy
& 0.264 & 0.191
& 0.274 & 0.189 & 0.268 & 0.192
& 0.242 & 0.172 & 0.252 & 0.184
& 0.052 & 0.059 & 0.053 & 0.051 \\
& 11n 100\% foggy
& 0.322 & 0.210
& 0.336 & 0.342 & 0.332 & 0.341
& 0.325 & 0.209 & 0.312 & 0.208
& 0.078 & 0.063 & 0.050 & 0.065 \\
& 11s clean
& 0.050 & 0.174
& 0.273 & 0.174 & -0.068 & 0.156
& 0.258 & 0.169 & -0.172 & 0.142
& 0.007 & 0.014 & -0.157 & 0.069 \\
& 11s 50\% foggy
& 0.306 & 0.194
& 0.370 & 0.240 & 0.336 & 0.195
& 0.351 & 0.204 & 0.321 & 0.187
& 0.157 & 0.149 & 0.155 & 0.124 \\
& 11s 100\% foggy
& 0.292 & 0.188
& 0.316 & 0.189 & 0.296 & 0.187
& 0.364 & 0.181 & 0.303 & 0.189
& 0.166 & 0.119 & 0.184 & 0.156 \\
& 11m clean
& -0.225 & 0.138
& 0.158 & 0.176 & 0.096 & 0.164
& 0.111 & 0.108 & 0.001 & 0.150
& 0.019 & 0.034 & 0.037 & 0.079 \\
& 11m 50\% foggy
& 0.302 & 0.207
& 0.311 & 0.207 & 0.303 & 0.205
& 0.339 & 0.258 & 0.277 & 0.191
& 0.079 & 0.078 & 0.051 & 0.069 \\
& 11m 100\% foggy
& 0.260 & 0.205
& 0.233 & 0.209 & 0.267 & 0.204
& 0.163 & 0.188 & 0.242 & 0.195
& 0.038 & 0.044 & 0.049 & 0.060 \\
\bottomrule

\end{tabular}%
}
\end{table}

\section{Discussion}

The results show that image restoration quality and downstream perception quality are not equivalent objectives. Although dehazing often improves visual clarity, the resulting gains in detection and tracking are not uniform across metrics, fog severities, or training settings. In the clean-trained case, restoration generally improves recall and mAP relative to foggy inputs, indicating partial recovery of target visibility and fewer missed detections. However, the restored results usually remain below the clean baseline, suggesting that restoration reduces, but does not remove, the clean-to-fog domain gap.

A likely reason is that restoration errors are spatially structured. Local contrast changes, edge distortions, or halo-like artifacts can alter the small image regions on which long-range UAV detection depends. Since the target occupies few pixels in sky-dominant scenes, even localized artifacts can affect detector confidence and bounding-box localization, which helps explain why visual enhancement does not necessarily translate into proportional gains in mAP@0.50:0.95 or tracking performance.

The experiments further indicate that robustness is driven more strongly by training-time adaptation than by restoration alone. Once foggy images are included in training, the performance gap between foggy and restored inputs becomes much smaller and the severe degradation observed for clean-trained detectors is substantially reduced. This shows that fog-inclusive training is the more reliable route to adverse-weather robustness, while restoration provides a conditional benefit that is most useful when the detector has not already adapted to fog.

The tracking results follow the same pattern. Across both ByteTrack and BoT-SORT, performance under fog is governed mainly by the robustness of the underlying detector. When fog causes missed detections, tracking degrades primarily through increased false negatives, and the tracker cannot fully recover once the target response is lost. Accordingly, fog-aware detector training transfers directly to stronger tracking performance, whereas the difference between ByteTrack and BoT-SORT remains secondary and sequence-dependent.

From a deployment perspective, restoration can be useful as a test-time preprocessing step for clean-trained detectors, especially under moderate and severe fog, but it adds computation and may introduce artifacts. Fog-inclusive training provides a more stable solution, although performance still degrades in the most challenging conditions. A practical system may therefore benefit from condition-aware operation, in which restoration or detector selection is adapted to the estimated visibility level and computational budget.

Finally, the present study should be interpreted within the scope of controlled synthetic-fog evaluation. The results provide a systematic analysis of relative robustness under physically motivated degradation, but they do not by themselves establish performance under real fog. Validation on real adverse-weather anti-UAV data remains an important direction for future work.

\section{Conclusions}
This paper presented a task-driven evaluation framework to study image restoration for UAV detection and tracking under synthetic fog. Using a depth-aware fog synthesis pipeline based on MiDaS depth estimation and the atmospheric scattering model, controlled fog severities were generated for sky-dominant, long-range imagery, enabling systematic analysis of how adverse visibility affects downstream perception. Within this framework, the study examined both detection and tracking performance and compared two complementary robustness strategies, namely test-time restoration and fog-inclusive detector training.

The results show that fog substantially degrades detection and tracking performance for small airborne targets, with the main failure mode arising from increased missed detections under severe degradation. Fog-inclusive training provides the most consistent improvement in robustness, reducing both the performance drop and its sensitivity to fog severity. Test-time restoration is most beneficial when the detector is trained only on clean imagery, where it can partially reduce the clean-to-fog domain gap. However, its effect becomes smaller and less consistent once the detector has already adapted to fog during training. These findings indicate that improvements in image-level restoration quality do not necessarily translate into proportional gains in downstream perception and should therefore be assessed jointly with detection and tracking metrics.

The tracking experiments further show that adverse-weather tracking performance is governed primarily by detector robustness rather than tracker choice alone. Both ByteTrack and BoT-SORT follow similar trends across visibility conditions, while fog-aware detector training transfers directly to stronger tracking-by-detection performance. From a practical perspective, the results suggest that robust deployment may benefit from condition-aware operation, in which detector selection or restoration is adapted to the estimated visibility level and computational constraints. Restoration may also remain useful in human-in-the-loop settings, where improved visual clarity can support operator awareness under degraded visibility.

The present study is limited to controlled synthetic-fog evaluation. Future work will extend the analysis to real adverse-weather data and to additional degradation types, including real fog, rain, and low-light conditions. It will also investigate real-time detection and tracking on edge devices, with and without image restoration, in order to better assess the practical trade-off between robustness and computational cost. A further direction is to study multimodal perception, for example camera--radar fusion, to determine how complementary sensing can improve UAV detection and tracking under adverse weather.

\vspace{6pt}

\section*{Author Contributions}
Conceptualization, A.P.; methodology, A.P.; software, A.P.; validation, A.P., V.A., and H.L.; formal analysis, A.P.; investigation, A.P., V.A., and H.L.; data curation, A.P.; resources, A.P., A.S., and H.N.; writing--original draft preparation, A.P.; writing--review and editing, A.P., V.A., H.L., A.S., and H.N.; visualization, A.P.; supervision, A.S. and H.N.; project administration, A.S. and H.N.; funding acquisition, A.S. and H.N. All authors have read and approved the manuscript.

\section*{Data Availability}
The datasets used in this study are publicly available from their original sources. The DUT Anti-UAV dataset is available at \url{https://github.com/wangdongdut/DUT-Anti-UAV} \cite{zhao2022vision}. The MMAUD dataset is available at \url{https://github.com/ntu-aris/MMAUD} \cite{yuan2024mmaud}. The spatially and temporally aligned visible and infrared UAV dataset is available at \url{https://doi.org/10.17632/sn9vy5c8sm.1} \cite{pereira2024infrared}. The synthetically generated foggy versions of the datasets used in this study are available at \url{https://doi.org/10.5281/zenodo.20615750}.

\section*{Conflicts of Interest}
The authors declare no conflicts of interest.

\section*{Abbreviations}
The following abbreviations are used in this manuscript:

\begin{center}
\begin{tabular}{@{}ll@{}}
\toprule
Abbreviation & Definition \\
\midrule
ASM & Atmospheric Scattering Model \\
CAP & Color Attenuation Prior \\
CNN & Convolutional Neural Network \\
DCP & Dark Channel Prior \\
FN & False Negative \\
FP & False Positive \\
IDF1 & Identity F1 Score \\
IoU & Intersection over Union \\
mAP & Mean Average Precision \\
MOTA & Multiple Object Tracking Accuracy \\
PSNR & Peak Signal-to-Noise Ratio \\
SSIM & Structural Similarity Index \\
TP & True Positive \\
UAV & Unmanned Aerial Vehicle \\
\bottomrule
\end{tabular}
\end{center}

\bibliographystyle{unsrtnat}
\bibliography{references}

\end{document}